# A Construction of Bayesian Networks from Databases Based on an MDL Principle

Joe Suzuki
Dept. of Industrial System Engineering
College of Sci. and Eng., Aoyama Gakuin University
6-16-1 Chitosedai, Setagaya-ku, Tokyo 157, Japan
e-mail: suzuki@ise.aoyama.ac.jp

## Abstract

This paper addresses learning stochastic rules especially on an inter-attribute relation based on a Minimum Description Length (MDL) principle with a finite number of examples, assuming an application to the design of intelligent relational database systems. The stochastic rule in this paper consists of a model giving the structure like the dependencies of a Bayesian Belief Network (BBN) and some stochastic parameters each indicating a conditional probability of an attribute value given the state determined by the other attributes' values in the same record. Especially, we propose the extended version of the algorithm of Chow and Liu in that our learning algorithm selects the model in the range where the dependencies among the attributes are represented by some general plural number of trees.

## 1 INTRODUCTION

This paper's main objective is to propose the algorithm learning a stochastic rule with examples based on an MDL (Minimum Description Length) principle with an application to the design of intelligent relational database systems that has been extensively reported [1,2,3], where the sequence of the examples is emitted by the stochastic rule, and each example is represented as a vector consisting of $R$ attributes, called an attribute vector [4,5,6,7].

Consider the situation where we design the intelligent relational database system inferring some missing attribute values from the other known values in the same record by using an available rule [8,9]. Then, we need to acquire some rules on the inter-attribute relation by some means first. However, the task of constructing such rules manually by an expert or with an expert [10] is very difficult and time-consuming [6]. So, we need the method that automatically learns correct rules with some given examples.

In this paper, we define learning stochastic rules on the inter-attribute relation as estimating the probability $P(x^R)$ of each $R$-dimensional attribute vector $x^R$ that consists of the model $g$ giving a structure such as a Bayesian Belief Network (BBN) [11,12] as well as the $k(g)$ stochastic parameters $p^{k(g)}$ each indicating the probability imbedded in the model such as a conditional probability, with the $n$ examples $x^R[n] = x_1^R x_2^R \cdots x_n^R$ of $R$-dimensional attribute vector $x_i^R$, $i = 1, 2, \cdots, n$. We assume that the $R$-dimensional attribute vector is represented as $x^R = (x^{(1)}, x^{(2)}, \cdots, x^{(R)})$, each attribute $x^{(j)}$ taking $\alpha_j$ values $j = 1, 2, \cdots, R$.

Much work is recently being devoted to learning stochastic rules with the $n$ examples $x^R[n]$ assuming arbitrary BBNs [11,12] as a model set $G$ [5,6,7,13,14,15]. BBN is represented as a directed acyclic graph in which nodes represent attributes and arcs between nodes represent probabilistic dependencies between the attributes, and is also the graphical tool that facilitates the qualitative structuring of uncertain knowledge and provides a framework for the numerical encoding of probabilistic relations. Especially, Cooper and Herskovits [6] have proposed the algorithm based on Bayesian, namely, selecting the model that maximizes the posterior probability $p(g|x^R[n])$ of the model $g \in G$ given the $n$ examples $x^R[n]$ while the other previous works are non-Bayesian.

This paper addresses the learning method based on the MDL (Minimum Description Length) principle [16,17], which is similar to the result of Cooper and Herskovits [6] in some sense, but differs in that ours assumes a prior probabilities neither on each model $g$ over the model set $G$ nor on each set of the $k(g)$ parameters $p^{k(g)}$ over the parametric space $[0,1]^{k(g)}$ when the model is $g$. The MDL principle selects the model such that the description length $l(x^R[n])$ of the $n$ examples $x^R[n]$, namely, the sum of both the description length $l_G(g)$ of the model $g \in G$ and the description length $l_g(x^R[n])$ of the $n$ examples $x^R[n]$ based on the model $g$ is minimized, preparing the description method for both of them beforehand. MDL selects the model achieving the best compromise between the simplicity of a selected model itself and the examples' fitnesses to the selected model, based on the description length as

$$l(x^R[n]) = \min_{g \in G}[l_G(g) + l_g(x^R[n])] \ . \quad (1)$$

In BBNs, for example, as the number of the edges in the network increases, the examples' fitnesses to the network are improved although the network gets more complex, having more conditional probabilities. However, as the number of the edges is reduced, the examples' fitnesses to the network get worse although the network has less conditional probabilities.

Our specific point is to apply the specific length function [18] $l(x^R[n])$ of the $n$ examples $x^R[n]$ achieving L. D. Davisson's minimax redundancy [19] as

$$\min_{l \in K} \max_{\theta \in \Lambda} \{ E_\theta[l(x^R[n])]/n - H(\theta) \} \ , \quad (2)$$

where $K$ is the set of the length function $l$ satisfying the Kraft's inequality as

$$\sum_{x^R[n]} 2^{-l(x^R[n])} \leq 1 \ , \quad (3)$$

and $E_\theta[\cdot]$ and $H(\theta)$ denote the average value on the stochastic rule $\theta$ and the per-example entropy of the stochastic rule $\theta$, respectively, where we use the base two logarithm through this paper without loss of generality. As long as the length function $l$ satisfies the Kraft's inequality, we have the method uniquely decoding the original $n$ examples with the described sequence [20]. We assume in this paper that the target range $\Lambda$ consists of the model set $G$ of the model $g$ and the parametric space $[0,1]^{k(g)}$ of the $k(g)$ stochastic parameters $p^{k(g)}$ when the model is $g \in G$. By minimizing the worst redundancy $\max_{\theta \in \Lambda} \{ E_\theta[l(x^R[n])]/n - H(\theta) \}$ for each $n$, indicating the correctness of the learning algorithm in an information theoretical sense, the worst redundancy converges to zero uniformly all over the target range $\Lambda$.

As a result of introducing the MDL principle without assuming any a prior probabilities on each stochastic rule $\theta \in \Lambda$, we derive a simple formula of description length for comparing the models as

$$l(x^R[n]) = \mathcal{H}(x^R[n]|g) + \frac{k(g)}{2}\log n \ , \quad (4)$$

where $\mathcal{H}(x^R[n]|g)$ is the empirical entropy using the $n$ examples $x^R[n]$ when the model is $g$, $k(g)$ is the number of the stochastic parameters in the model $g$, and we assume that each model included in the model set $G$ is represented as state decomposition [21].

Furthermore, we show the extended version of a Chow and Liu algorithm [7] in the sense that our learning algorithm selects the model in the range where the dependencies among the attributes are represented by the general plural number of trees rather than one tree necessarily connecting all the attributes as the Chow and Liu algorithm [7]. Their algorithm selects the model that the Kullback-Leibler information is minimized between the true stochastic rule $\theta$ and the learned stochastic rule whose model is restricted to the set of the models representing a tree, by using Kruskal's minimum spanning tree algorithm [22] that minimizes the tree's sum of the cost, that is, the mutual information $I(X^{(i)}, X^{(j)})$ attached to the edge between two nodes $x^{(i)}$ and $x^{(j)}$, $i,j = 1,2,\cdots R^1$, where the nodes and the edges in the tree correspond to the attributes and the dependencies, respectively. The proposed algorithm connects the edge between the two nodes $x^{(i)}$ and $x^{(j)}$ if and only if the mutual information $I(X^{(i)}, X^{(j)})$ is more than $(\alpha_i - 1)(\alpha_j - 1) \log n/2n$.

## 2 DISCUSSION WITHOUT ASSUMING BAYESIAN BELIEF NETWORKS

First, let us discuss a few basic results on learning stochastic rules on the input-output relation based on the MDL principle before developing the results restricted to the rule on the inter-attribute relation or the BBNs.

### 2.1 LEARNING STOCHASTIC RULES REPRESENTED AS INPUT-OUTPUT RELATION

Consider the problem of learning stochastic rules represented on the input-output relation with the $n$ examples

$$\begin{aligned} x_1^N &= (x_1^{(1)}, x_1^{(2)}, \cdots, x_1^{(N)}) \quad , \quad y_1 \\ x_2^N &= (x_2^{(1)}, x_2^{(2)}, \cdots, x_2^{(N)}) \quad , \quad y_2 \\ &\cdots \cdots \\ x_n^N &= (x_n^{(1)}, x_n^{(2)}, \cdots, x_n^{(N)}) \quad , \quad y_n \ , \end{aligned}$$

where $x_i^{(j)} \in A_j = \{0, 1, \cdots, \alpha_j - 1\}$, $j = 1, 2, \cdots, N$, $y_i \in A_{N+1} = \{0, 1, \cdots, \alpha_{N+1} - 1\}$, $i = 1, 2, \cdots, n$. We call $y_i$ a class instead of an attribute value. We define this problem as estimating the conditional probability $P(y|x^N)$ of the class $y$ given the attribute vector $x^N$ in terms of the model $g$ as well as the $k(g)$ stochastic parameters $p^{k(g)}$.

We represent each model $g \in G$ as the way of decomposing the $N$-dimensional space $\prod_{j=1}^{N} A_j$ of the attribute vector $x_i^N$ into $S(g)$ states. That is, we divide the $N$-dimensional product space $\prod_{j=1}^{N} A_j$ into the $S(g)$ sets $B_s$, $s = 1, 2, \cdots, S(g)$ including some $N$-dimensional vectors: we have $\cup_{s=1}^{S(g)} B_s = \prod_{j=1}^{N} A_j$ and $B_s \cap B_{s'} = \phi$ for any $s \neq s'$, where we call each label $s = 1, 2, \cdots, S(g)$ of the set $B_s$ a state [21]. Then we can define the probability $p[q, s, g]$ of the class $q \in A_{N+1}$ given the state $s = 1, 2, \cdots, S(g)$ when the model is $g$.

---

[1] We denote a stochastic variable and a real value as an upper case and a lower case, respectively.

268  Suzuki

**Example 1** *Suppose $N = 2$ and $\alpha_1 = \alpha_2 = 2$. The problem is to divide the four vectors $x^2 = (x^{(1)}, x^{(2)}) = (0,0), (0,1), (1,0), (1,1)$ into some number of the states. We can make at most the fifteen models as in Table 1:*

Table 1: Assignment of $x^2$ to $s$ in Example 1

| MODEL $g$ | # of STATES $S(g)$ | $x^2 = (x^{(1)}, x^{(2)})$ | | | |
|---|---|---|---|---|---|
| | | (0,0) | (0,1) | (1,0) | (1,1) |
| 1 | 1 | s=1 | s=1 | s=1 | s=1 |
| 2 | 2 | s=1 | s=1 | s=1 | s=2 |
| 3 | 2 | s=1 | s=1 | s=2 | s=1 |
| 4 | 2 | s=1 | s=1 | s=2 | s=2 |
| 5 | 2 | s=1 | s=2 | s=1 | s=1 |
| 6 | 2 | s=1 | s=2 | s=1 | s=2 |
| 7 | 2 | s=1 | s=2 | s=2 | s=1 |
| 8 | 2 | s=1 | s=2 | s=2 | s=2 |
| 9 | 3 | s=1 | s=1 | s=2 | s=3 |
| 10 | 3 | s=1 | s=2 | s=1 | s=3 |
| 11 | 3 | s=1 | s=2 | s=3 | s=1 |
| 12 | 3 | s=1 | s=2 | s=2 | s=3 |
| 13 | 3 | s=1 | s=2 | s=3 | s=2 |
| 14 | 3 | s=1 | s=2 | s=3 | s=3 |
| 15 | 4 | s=1 | s=2 | s=3 | s=4 |

We call this representational method state decomposition for the input-output relation in this paper although the original concept of this framework is much wider as in [21].

Although there exists a finite number of the models in the model set $G$ represented as the state decomposition in this paper because the combination of the states is finite, however, such a combination is extraordinarily large. So, we need to restrict the model set $G$ when we actually use state decomposition.

**Theorem 1** *The number of the models that we can represent using the state decomposition for the input-output relation is $f[\prod_{j=1}^{N} \alpha_j]$ when the $j$-th attribute and the class take $\alpha_j$ values, $j = 1, 2, \cdots, N$ and $\alpha_{N+1}$ values, respectively, where the function $f[\cdot]$ is defined as*

$$f[m] = \sum_{S=1}^{m} \sum_{T=1}^{S} \frac{T^m (-1)^{S-T}}{(S-T)!T!} . \quad (5)$$

This result is the modified version of [4] assuming $\alpha_j = \alpha$, where $\alpha \geq 2$ is an integer. By using this result, we can compute as fifteen, the combination of the models that we can represent using the state decomposition in Example 1.

In general, the length function $l_g(y[n]|x^N[n])$ of the sequence of the class $y[n]$ given the sequence of the attribute vector $x^N[n]$ when the model $g$ is fixed can be written as

$$-\log\{\sum_{p^{k(g)} \in [0,1]^{k(g)}} w(p^{k(g)}) P[p^{k(g)}](y[n]|x^N[n])\} , \quad (6)$$

where

$$k(g) = S(g)(\alpha_{N+1} - 1) , \quad (7)$$

$P[p^{k(g)}](y[n]|x^N[n])$ is the probability of $y[n]$ given $x^N[n]$ when the $k(g)$ parameters are $p^{k(g)}$, and the function $w(\cdot)$, called a weight function [23,24], must satisfy the inequality as

$$\sum_{p^{k(g)} \in [0,1]^{k(g)}} w(p^{k(g)}) \leq 1 . \quad (8)$$

The number $k(g)$ of the stochastic parameters $p^{k(g)}$ is computed by using Eq. (7) because the one probability of the $\alpha_{N+1}$ classes is calculated with the other $\alpha_{N+1} - 1$ probabilities belonging to the same state. Also, the value of $k(g)$ is derived when we calculate the description length as we will see in Theorem 2. So, the problem of determining the weight function is reduced to setting the weight function. Let the class of the weight function be the Dirichlet distribution with one parameter $a > 0$, namely,

$$w(p^{k(g)}) = \prod_{s=1}^{S(g)} \{\frac{\Gamma(\alpha_{N+1}a)}{[\Gamma(a)]^{\alpha_{N+1}}} \prod_{q=0}^{\alpha_{N+1}-1} p[q, s, g]^{a-1}\} , \quad (9)$$

where $\Gamma[x]$ is the gamma function of $x$ as

$$\Gamma(x) = \int_0^\infty z^{x-1} e^{-z} dz . \quad (10)$$

Then, by choosing $a = 1/2$ rather than $a = 1$ assuming that the weight function has the uniform distribution proposed by Cooper and Herskovits [6], we have the length function achieving the minimax redundancy except for the terms of $O(1/n)$ [21,25,26,23,24]. In other words, there exists a constant $C_0$ satisfying

$$E_\theta[l_g(y[n]|x^N[n])]/n - H(\theta|x^N[n]) \leq \frac{k(g_\theta)}{2n} \log n + \frac{C_0}{n} , \quad (11)$$

where $H(\theta|x^N[n])$ is the entropy of the stochastic rule $\theta$ given the input $x^N[n]$, and $g_\theta$ is the true model of the stochastic rule $\theta$.

**Theorem 2** *Choosing one parameter $a > 0$ in Eq. (9) as $1/2$, the length function $l_g(y[n]|x^N[n])$ in Eq. (6) is reduced to*

$$l_g(y[n]|x^N[n]) = \mathcal{H}(y[n]|x^N[n], g) + \frac{k(g)}{2} \log n \quad (12)$$

*except for the terms of $O(1)$, where the empirical entropy $\mathcal{H}(y[n]|x^N[n], g)$ is defined as*

$$\mathcal{H}(y[n]|x^N[n], g) = \sum_{s=1}^{S(g)} \sum_{q=0}^{\alpha_{N+1}-1} -n[q, s, g] \log \frac{n[q, s, g]}{n[s, g]} , \quad (13)$$

*and $n[q, s, g]$ and $n[s, g]$ are respectively the occurrence of the class $q$ on the state $s$ and the occurrence of the state $s$ when the model is $g$.*

(For the proof of Theorem 2, see Appendix A.)

Then we have (11) because

$$E_\theta[\mathcal{H}(y[n]|x^N[n],g)] = nH(\theta|x^N[n]) + O(1) \quad (14)$$

holds.

Notice that no result has been reported that the length function letting $a$ be one satisfies the property in Theorem 2. Futhermore, in a similar case, we have shown that the lower bound of the minimax redundancy coincides with the RHS of Eq. (11) except for the terms of $O(1/n)$ [23,24]. We can derive the lower bound of RHS of Eq. (11) similarly.

On the other hand, the description length $l_G(g)$ of the model $g \in G$ itself also must satisfy the Kraft's inequality as

$$\sum_{g \in G} 2^{-l_G(g)} \leq 1 \ . \quad (15)$$

In order to minimize the worst redundancy, we must set the description length $l_G(g)$ of the model $g \in G$ as

$$l_G(g) = 1/|G| \ . \quad (16)$$

**Theorem 3** *The worst redundancy*

$$\max_{\theta \in \Lambda}\{E_\theta[l(y|x^N[n])]/n - H(\theta|x^N[n])\}$$

*is upper-bounded by*

$$\frac{k(g_\theta)}{2n}\log n + \frac{C_1}{n} \ , \quad (17)$$

*for any stochastic rule whose model set is $G$, where $C_1 = C_0 + \log |G|$.*

### 2.2 Learning Stochastic Rules Represented as Inter-Attribute Relation

Consider the problem of learning stochastic rules on the inter-attribute relation with $n$ examples

$$x_1^R = (x_1^{(1)}, x_1^{(2)}, \cdots, x_1^{(R)})$$
$$x_2^R = (x_2^{(1)}, x_2^{(2)}, \cdots, x_2^{(R)})$$
$$\cdots \cdots$$
$$x_n^R = (x_n^{(1)}, x_n^{(2)}, \cdots, x_n^{(R)}) \quad ,$$

where $x_i^{(j)} \in A_j = \{0, 1, \cdots, \alpha_j - 1\}$, $j = 1, 2, \cdots, R$, $i = 1, 2, \cdots, n$. We define this problem as estimating the probability $P(x^R)$ of the attribute vector $x^R$ in terms of the model $g$ as well as the $k(g)$ stochastic parameters $p^{k(g)}$.

We obtain the solution by iterating the model selection procedure similar to the case of the input-output relation for $N = 1, 2, \cdots, R-1$, where each model $g$ in the model set $G$ decomposes the $N$-dimensional space $\prod_{j=1}^{N} A_j$ of the attribute vector $x^N$ into $S_N(g)$ states at each $N$-th stage, after setting the one state at the initial stage, namely, $S_0(g) = 1$ for any $g \in G$. This is because we can describe the probability $P(x^R)$ of the attribute vector $x^R$ as

$$P(x^R) = p(x^{(1)})p(x^{(2)}|x^{(1)}) \cdots p(x^{(R)}|x^{(1)}x^{(2)} \cdots x^{(R-1)}) \ . \quad (18)$$

Intuitively, at the $N$-th stage, we can regard the $N$-dimensional attribute vector

$$x^N = (x^{(1)}, x^{(2)}, \cdots, x^{(N)})$$

and the one attribute $x^{(N+1)}$, $N = 0, 1, \ldots, R-1$, as the attribute vector and the class, respectively, in learning stochastic rules represented as the input-output relation.

**Theorem 4** *The number of the models that we can represent using the state decomposition for the inter-attribute relation and the number of the comparison required for selecting one model $g \in G$ are respectively $\prod_{N=0}^{R-1} f[\prod_{j=1}^{N} \alpha_j]$ and $\sum_{N=0}^{R-1} f[\prod_{j=1}^{N} \alpha_j] - R$ when the $j$-th attribute takes $\alpha_j$ values $j = 1, 2, \cdots, R$.*

This result is derived from Theorem 1 straightforward.

**Theorem 5** *We have the length function in Eq. (4) satisfying the Kraft's inequality, except for the terms of $O(1)$, where the empirical entropy $\mathcal{H}(x^R[n]|g)$ is defined as*

$$\mathcal{H}(x^N[n]|g) = \sum_{N=0}^{R-1} \sum_{s=1}^{S_N(g)} \sum_{q=0}^{\alpha_{N+1}-1} -n[q,s,N,g]\log\frac{n[q,s,N,g]}{n[s,N,g]} \quad (19)$$

*and $n[q, s, N, g]$ and $n[s, N, g]$ are respectively the occurrence of the class $q$ on the state $s$ and the occurrence of the state $s$ at the $N$-th stage when the model is $g$, and*

$$k(g) = \sum_{N=0}^{R-1} S_N(g)(\alpha_{N+1} - 1) \ . \quad (20)$$

Similar to the input-output case, there exists a constant $C_2$ satisfying

$$E_\theta[l_g(x^R[n])]/n - H(\theta) \leq \frac{k(g_\theta)}{2n}\log n + \frac{C_2}{n} \ . \quad (21)$$

**Theorem 6** *The worst redundancy*

$$\max_{\theta \in \Lambda}\{E_\theta[l(x^R[n])]/n - H(\theta)\}$$

*is upper-bounded by*

$$\frac{k(g_\theta)}{2n}\log n + \frac{C_3}{n} \ , \quad (22)$$

*for any stochastic rule whose model set is $G$ where $C_3 = C_2 + \log |G|$.*

## 3 DISCUSSION ASSUMING BAYESIAN BELIEF NETWORKS

### 3.1 GENERAL CASE

Assuming arbitrary BBNs amounts to restricting the model class $G$ in the state decomposition to a specific case. In BBNs, we just need to decide for any two nodes of the network whether we connect the edge between them or not. Without loss of generality, suppose that we connect the edges from each element of the parent set $\pi^N$ [6] to the node $x^{(N+1)}$, $N = 1, 2, \cdots, R$, where no edge is connected if the parent set $\pi^N$ is empty. In other words, we can describe the probability $P(x^R)$ of the atribute vector $x^R$ as

$$P(x^R) = p(x^{(1)}|\pi^0)p(x^{(2)}|\pi^1) \ldots p(x^{(R)}|\pi^{R-1}) , \quad (23)$$

where the parent set $\pi^0$ of the node label 0 is empty. We can exchange the node labels $N = 1, 2, \cdots, R$ so that Eq. (23) holds since BBN is acyclic and has no loops of the directed edges in the network [11,12].

When we compute the conditional probability $P(x^{(N+1)}|\pi^N)$, we divide the space of the attributes included in $\pi^N$ into the $S_N(g)$ states, where the number of the way of decomposing the states is $2^N$ rather than $f[\prod_{j=1}^N \alpha_j]$. Let the set of each label corresponding to the attribute included in $\pi^N$ be $\phi^N$. Then we can divide the vector consisting of the attributes $x^{(j)}$, $j \in \phi^N$ into the $S_N(g)$ states.

**Theorem 7** *In BBNs, the number of the states at the N-th state when the model is g is computed as*

$$S_N(g) = \prod_{j \in \phi^N} \alpha_j . \quad (24)$$

*Therefore,*

$$k(g) = \sum_{N=0}^{R-1} (\alpha_{N+1} - 1) \prod_{j \in \phi^N} \alpha_j \quad (25)$$

*holds.*

Then, we can select the model using Eqs.(4), (13) and (25) so that the description length $l(x^R[n])$ is minimized. Cooper and Herskovits [6] have proposed two methods for finding the model that maximizes the posterior probability $P(g|x^R[n])$ of the model $g$ given the $n$ examples $x^R[n]$: first, the one specifying the optimal model by comparing the $2^{\binom{R}{2}}$ models; second, the polynominal-time heuristic method adding each node to the parent set incrementally until the performance of the resulting model is not improved, without insuring the optimality.

On the other hand, we can also apply our proposed method to the above two search strategies. We can continue adding the edge to the network while the amount reducing the empirical entropy indicating the fitness of the $n$ examples $x^R[n]$ to the network is larger than the increase in the complexity of the model $g$ proportional to the number $k(g)$ of the stochastic parameters $p^{k(g)}$. Anyway, we need the model selection to be global rather than sequential when we want to insure the optimality at the expense of computational efforts.

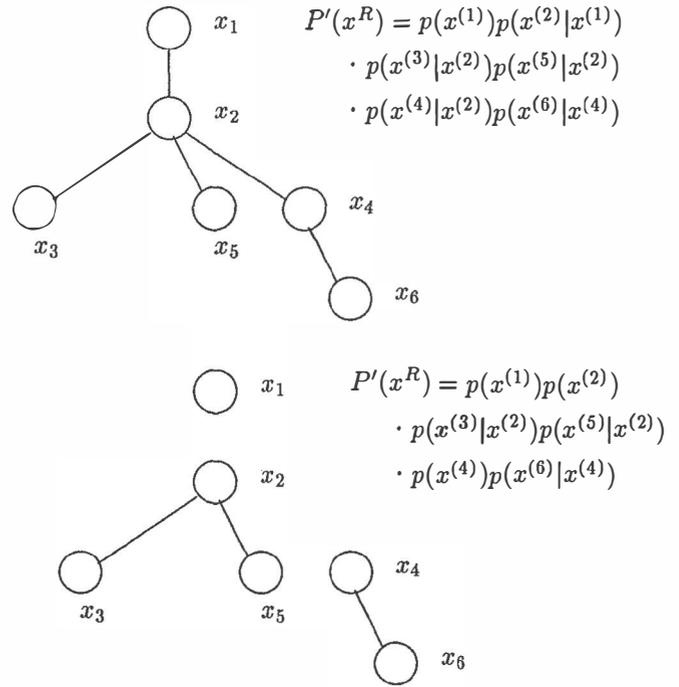

Figure 1: An Example of A Tree

### 3.2 DISCUSSION ASSUMING TREES

The merit of our proposed method compared to the result of Cooper and Herskovits [6] is not so clear even through the discussion of the above two search strategies, except that ours needs no a prior probability on each stochastic rule. In this subsection, however, we show that learning stochastic rules based on the MDL principle leads our method to the interesting and elegant algorithm extending the Chow and Liu algorithm [7] when we assume that the network consists of trees, where we need not connect all the nodes through the path of any edges each other unlike Chow and Liu [7].

**Assumption 1** *We allow the way of decomposing some states to depend on at most one attribute $x^{q[N]}$, $1 \leq q[N] \leq N$ at the N-th stage, $N = 0, 1, \ldots, R-1$ when the model is $g \in G$, where $q[0] = 0$, and $q[N] = 0$ iff the parent set $\phi^N$ of the node $x^N$ is empty.*

That is, we learn the stochastic rules among the probabilities in terms of

$$P'(x^R) = \prod_{N=1}^R p(x^{(N+1)}|x^{(q[N])}) , \quad 0 \leq q[N] \leq N ,$$
(26)

called a Dendroid distribution, rather than the probability in Eq. (18), where we can consider the $R!$ models according to the value of $0 \leq q[N] \leq N$ for each $N = 0, \cdots, R-1$. Chow and Liu [7] have proposed a minimum cost spanning tree [22] whose cost of the edge between two nodes $x^{(i)}$ and $x^{(j)}$ is the value of the mutual information $I(X^{(i)}, X^{(j)})$. The resulting Dendroid distribution minimizes Kullback-Leibler's information [27] between $P(x^R)$ and $P'(x^R)$ in the range $q[N] \neq 0$, $N = 1, 2, \ldots, R - 1$. In other words, any two nodes must be connected through any path of the edges.

Let us first show the proposed algorithm. In Algorithm 1, a priority queue $Q$ is working memory, thereby a set $T$ is the output.

### Algorithm 1
**begin**

1. *the set $T := 0$;*

2. *Compute the mutual information $I(X^{(i)}, X^{(j)})$ for all the edge $(i, j)$; sort the edges in the descending order of $I(X^{(i)}, X^{(j)}) - (\alpha_i - 1)(\alpha_j - 1) \log n / 2n$, and store them into the priority queue $Q$;*

3. *Let the set of the sets $\{j\}$ $j = 1, 2, \ldots, R$ be $VS$;*

4. **while** *the maximum value in $Q$ of $I(X^{(i)}, X^{(j)}) - (\alpha_i - 1)(\alpha_j - 1) \log n / 2n > 0$* **do**
   **begin**

   *(a) Remove such an edge $(i, j)$ that maximizes $I(X^{(i)}, X^{(j)}) - (\alpha_i - 1)(\alpha_j - 1) \log n / 2n$ from $Q$;*

   *(b)* **if** *The two label $i$ and $j$ are included in the different sets in $VS$ such as a set $W_1$ and a set $W_2$* **then**
   **begin**
   *Replace $W_1$ and $W_2$ in $VS$ for $W_1 \cup W_2$; add $(i, j)$ to $T$*
   **end**

   **end**

**end.**

In Algorithm 1, the mutual information $I(X^{(i)}, X^{(j)})$ is defined as

$$I(X^{(i)}, X^{(j)}) = -\sum_{x^{(i)}, x^{(j)}} \hat{p}(x^{(i)}, x^{(j)}) \log \frac{\hat{p}(x^{(i)})}{\hat{p}(x^{(i)}|x^{(j)})} \quad (27)$$

and $\hat{p}(\cdot)$ (or $\hat{p}(\cdot, \cdot)$) denotes the maximum likelihood estimator obtained with the relative occurrence of the $n$ examples.

If the mutual information attached to that edge between two nodes $x^{(i)}$ and $x^{(j)}$ is larger than $(\alpha_i - 1)(\alpha_j - 1) \log n / 2n$ and connecting them yields to making no loops, we can reduce the description length by adding it to the tree.

**Theorem 8** *Algorithm 1 selects the set of trees minimizing the description length of the n examples $x^R[n]$ among the Dendroid distributions represented as Eq. (26).*

(For the proof of Theorem 8, see Appendix B.)

On the other hand, we can replace the term

$$(\alpha_i - 1)(\alpha_j - 1) \log n / 2n \quad (28)$$

in Algorithm 1 for the general term of the function $c(n)$ of $n$ as

$$(\alpha_i - 1)(\alpha_j - 1) c(n) / 2 , \quad (29)$$

where MDL and AIC (Akaike's Information Criterion) [28] correspond to the case $c(n) = \log n$ and $c(n) = 2$, respectively. Furthermore, we might say that the result of Chow and Liu corresponds to the case $c(n) = 0$, thereby that their algorithm selects the maximum likelihood model.

## 4 Concluding Remarks

We present the various learning algorithms as:

- the rules represented as the general BBNs;
- the rules represented as BBNs consisting of the general plural trees; as well as
- the rules represented as the general input-output and inter-relation relation without assuming any BBNs

based on the MDL principle with the application to the design of the intelligent relational database systems.

Our future topics concerning this paper include:

1. problem-solving in actual database designs: we are now planning the design of a management relational database in detail; and

2. the reasonable way of setting the $k(g)$ stochastic parameters: we recommend estimating the value of $p[q, s, N, g]$ as

$$p[q, s, N, g] = \frac{n[q, s, N, g] + a}{n[s, N, g] + a\alpha_{N+1}} \quad (30)$$

where $a = 1/2$, because the length functions in this paper are derived using $a = 1/2$. However, we do not fully know the property of this parameter estimator.

### Acknowledgements

Acknowledgements: The author would like to thank Prof. Shigeichi Hirasawa of Waseda University and Mr. Yasutaka Ohdake of Toshiba corporation for his fruitful discussions and comments. This work was partially supported by the Telecommunications Advancement Foundation (TAF).

### References

[1] W.J. Frawley. Knowledge discovery in databases: an overview. In *Knowledge Discovery in Databases edited by G.P. Shapiro and W.J. Frawley*, pages 1–27, 1991.

[2] K.C.C.Chan and A.K.C. Wong. A statistical technique for extracting classificatory knowledge from database. In *Knowledge Discovery in Databases edited by G.P. Shapiro and W.J. Frawley*, pages 107–124, 1991.

[3] J. Ullman. Database systems: achievements and opportunities. *Communications of ACM*, Oct. 1991.

[4] J. Suzuki, Y. Ohdake, and S. Hirasawa. On the selection method of stochastic models based on a minimum description length principle and a state decomposition. *IEICE Tech. Rep.*, IT90-105, Feb. 1991. (in Japanese).

[5] T. S. Verma and J. Pearl. Equivalence and synthesis of causal models. In *Uncertainty in Artificial Intelligence '90*, pages 220–227, 1990.

[6] G.F. Cooper and E. Herskovits. A Bayesian method for constructing Bayesian belief networks from database. In *Uncertainty in Artificial Intelligence '91*, pages 86–94, 1991.

[7] C.K. Chow and C.N. Liu. Approximating discrete probability distributions with dependence trees. *IEEE Trans. on Inform. Theory*, IT-14(3):462–467, May 1968.

[8] Y. Huang, J. Han, N. Cercone, and G. Hall. Reasoning with inductive dependencies. In *AI and Statistics '93*, pages 283–288, 1993.

[9] W.X. Wen. From relational database to belief network. In *Uncertainty in Artificial Intelligence '92*, pages 406–413, 1992.

[10] D.E. Heckerman. *Probabilistic Similarity Networks*. PhD thesis, Medical Information Science, Stanford University, 1990.

[11] J. Peral. *Probabilistic Reasoning in Intelligent Systems*. Morgan Kaufmann, San Mateo, California, 1988.

[12] J. Pearl. Fusion, propagation and structuring in belief networks. *Artificial Intelligence*, 29:241–288, 1986.

[13] D. Geiger, A. Paz, and J. Pearl. Learning causal trees from dependence information. In *AAAI '90*, 1990.

[14] R.M. Fung and S.L. Crawford. Constructor: a system for the induction of probabilistic models. In *AAAI '90*, 1990.

[15] S. Srinivas, S. Russel, and A. Agogino. Automated construction of sparse bayesian networks for unstructured probabilistic models and domain information. In *M. Henrion, R.D. Shachter, L.N. Kanal and J.F. Lemmer (Eds), Uncertainty in Artificial Intelligence 5 (North-Holland, Amsterdam)*, 1990.

[16] J. Rissanen. Universal coding, information, prediction, and estimation. *IEEE Trans. on Inform. Theory*, IT-30(4):629–636, July 1984.

[17] J. Rissanen. Stochastic complexity and modeling. *The Annals of Statistics*, 14(3):1080–1100, 1986.

[18] J. Rissanen. A universal prior integer and estimation by minimum description length. *The Annals of Statistics*, 11(2):416–431, 1983.

[19] L.D. Davisson. Universal noiseless coding. *IEEE Trans. Inform. Theory*, IT-19(6):783–795, Nov. 1973.

[20] C.E. Shannon. A mathematical theory of communication. *Bell System Tech. Journal*, 27:379–423, 1948.

[21] J. Suzuki. Generalization of the learning method for classifying rules with consistency irrespective of the representation form and the number of the classified patterns. In *ISITA 90*, Waikiki, Hawaii, Nov. 1990.

[22] J.B. Kruskal Jr. On the shortest spanning subtree of a graph and the travelling salesman problem. In *Proc. Amer. Math. Soc. 7:1*, pages 48–50, 1956.

[23] J. Suzuki. Minimax redundancy for sources with an unknown model. In *IEEE ISIT'93*, page 58, Jan. 1993.

[24] J. Suzuki. Universal coding scheme minimizing minimax redundancy for sources with an unknown model. *Trans. IEICE, Part-A*, Aug. 1993. (to appear).

[25] L.D. Davisson, R.J. Mceliece, M.B. Pursley, and M.S. Wallace. Efficient universal noiseless source codes. *IEEE Trans. Inform. Theory*, IT-27(3):269–279, May 1981.

[26] L.D. Davisson. Minimax noiseless universal coding for Markov sources. *IEEE Trans. Inform. Theory*, IT-29(2):211–215, March 1983.

[27] S. Kullback. *Information Theory and Statics*. New York: John Wiley and Sons, London: Chapman and Hall, 1959.

[28] H. Akaike. A new look at the statistical model identification. *IEEE Trans. on Automatic Control*, AC-19(6):716–723, Dec. 1974.

# Appendix A  Proof of Theorem 2

Notice

$$
\begin{aligned}
& \int_0^1 w(p^{k(g)}) P(x_1^n|\theta) d\theta \\
&= \int_0^1 \prod_{s=1}^{S(g)} \frac{\Gamma(\alpha_{N+1}a)}{[\Gamma(a)]^{\alpha_{N+1}}} \prod_{q=0}^{\alpha_{N+1}-1} p[q,s,g]^{n[q,s,g]+a-1} dp^{k(g)} \\
&= \prod_{s=1}^{S(g)} \Gamma(\alpha_{N+1}a) \frac{\prod_{q=0}^{\alpha_{N+1}-1} \Gamma(n[q,s,g]+a)}{[\Gamma(a)]^{\alpha_{N+1}} \Gamma(n[s,g]+\alpha_{N+1}a)} \\
&= \prod_{s=1}^{S(g)} \frac{\prod_{q=0}^{\alpha_{N+1}-1} \{(n[q,s,g]+a-1)!\}}{(n[s,g]+\alpha_{N+1}a-1)!} . \quad\quad (A.1)
\end{aligned}
$$

Thereby, let $a$ be $1/2$ and evaluate the value of (A.1) by using Stirling's formula as

$$n! = \sqrt{2\pi} n^{n+1/2} \exp\{-n+\theta_n\}, \ 0 < \theta_n < \frac{1}{12n} \ . \quad (A.2)$$

In general, we have

$$(n + \frac{\alpha - 2}{2})!$$
$$= n! n^{\frac{\alpha-2}{2}} \prod_{i=1}^{\frac{\alpha-2}{2}} \frac{n+i}{n}$$
$$= n^{n+\frac{\alpha-1}{2}} \exp\{-n + \ln\sqrt{2\pi} + O(1/n)\} \quad (A.3)$$

for $n$ even and

$$(n + \frac{\alpha - 2}{2})!$$
$$= (n - \frac{1}{2})! n^{\frac{\alpha-1}{2}} \prod_{i=1}^{\frac{\alpha-1}{2}} \frac{n+i-\frac{1}{2}}{n}$$
$$= n^{n+\frac{\alpha-1}{2}} \exp\{-n + \ln\sqrt{2} + O(1/n)\} \quad (A.4)$$

for $n$ odd, where $\alpha$ is an integer no less than 2. Therefore,

$$-\log \prod_{s=1}^{S(g)} \frac{\prod_{q=0}^{\alpha_{N+1}-1}(n[q,s,g] - \frac{1}{2})!}{(n[s,g] + \alpha_{N+1} - \frac{1}{2})!}$$
$$= \sum_{s=1}^{S(g)} \sum_{q=0}^{\alpha_{N+1}-1} -n[q,s,g] \log \frac{n[q,s,g]}{n[s,g]}$$
$$+ \sum_{s=1}^{S(g)} \frac{\alpha_{N+1} - 1}{2} \log n[s,g] + O(1) \quad (A.5)$$

holds, where the value $\log n[s,g]$ is upper-bounded by $\log n$.

## Appendix B  Proof of Theorem 8

First, consider the case where we connect the node $x^{(N+1)}$ with the node $x^{(q[N])}$, $1 \le q[N] \le N$. We prepare the $S_N(g) = \alpha_{q[N]}$ states for the node $x^{(N+1)} = 0, 1, \cdots, \alpha_{N+1} - 1$. So, the number of the stochastic parameters associated with the node $x^{(N+1)}$ is

$$(\alpha_{N+1} - 1)\alpha_{q[N]} \ . \quad (B.6)$$

Also, since we have

$$\frac{n[q,s,N,g]}{n[s,N,g]} = \hat{p}(x^{(N+1)}|x^{(q[N])}) \quad (B.7)$$

and

$$\frac{n[q,s,N,g]}{n} = \hat{p}(x^{(N+1)}, x^{(q[N])}) \quad (B.8)$$

from the definition of $\hat{p}(\cdot|\cdot)$ and $\hat{p}(\cdot,\cdot)$,

$$\sum_{s=1}^{S_N(g)} \sum_{q=0}^{\alpha_{N+1}-1} -n[q,s,N,g] \log \frac{n[q,s,N,g]}{n[s,N,g]}$$

$$= n \sum_{x^{(N+1)}=0}^{\alpha_{N+1}-1} \sum_{x^{(q[N])}=0}^{\alpha_{N+1}-1} -\hat{p}(x^{(N+1)}, x^{(q[N])})$$
$$\cdot \log \hat{p}(x^{(N+1)}|x^{(q[N])})$$

$$= n \sum_{x^{(N+1)},x^{(q[N])}} \hat{p}(x^{(N+1)}, x^{(q[N])}) \cdot \log \frac{\hat{p}(x^{(N+1)})}{\hat{p}(x^{(N+1)}|x^{(q[N])})}$$
$$- n \sum_{x^{(N+1)}} \hat{p}(x^{(N+1)}) \log \hat{p}(x^{(N+1)})$$
(B.9)

$$= -n\hat{I}(X^{(N+1)}, X^{(q[N])}) + n\hat{H}(X^{(N+1)}) \quad (B.10)$$

holds, where

$$\hat{H}(X^{(N+1)}) = -\sum_{x^{(N+1)}} \hat{p}(x^{(N+1)}) \log \hat{p}(x^{(N+1)}) \ . \quad (B.11)$$

Next, consider the case where we do not connect the node $x^{(N+1)}$ with the node $x^{(q[N])}$, $q[N] = 0$. We prepare the $S_N(g) = 1$ state for the node $x^{(N+1)} = 0, 1, \cdots, \alpha_{N+1} - 1$. So the number of the stochastic parameters associated with the node $x^{(N+1)}$ is

$$\alpha_{N+1} - 1 \ . \quad (B.12)$$

On the other hand, letting $q[N]$ be zero in Eq. (B.9), we derive

$$\sum_{s=1}^{S_N(g)} \sum_{q=0}^{\alpha_{N+1}-1} -n[q,s,N,g] \log \frac{n[q,s,N,g]}{n[s,N,g]} = n\hat{H}(X^{(N+1)}) \ . \quad (B.13)$$

Therefore, we can compute the description length $l(x^R[n])$ as

$$l(x^R[n]) = -n \sum_{q[N] \ne 0} \hat{I}(X^{(N+1)}, X^{(q[N])})$$
$$+ n \sum_{N=0}^{R-1} \hat{H}(X^{(N+1)}) + \frac{k(g)}{2} \log n + C_2 \ , \quad (B.14)$$

where

$$k(g) = \sum_{N=0}^{R-1} (\alpha_{N+1} - 1)\alpha_{q[N]} \ , \quad (B.15)$$

and $\alpha_0 = 1$.

Since the value of the second term in Eq. (B.14) is constant in computing the description length, we just need to use the value of

$$-\hat{I}(X^{(N+1)}, X^{(q[N])}) + \frac{(\alpha_{N+1} - 1)(\alpha_{q[N]} - 1) \log n}{2n} \quad (B.16)$$

when we decide whether we connect the edge $(q[N], N+1)$ or not. This completes the proof.